%% file: arxivver.tex
\documentclass[10pt,twocolumn,letterpaper]{article}

\usepackage[pagenumbers]{cvpr} 

\usepackage{graphicx}
\usepackage{amsmath}
\usepackage{amssymb}
\usepackage{booktabs}
\usepackage{bbm}
\usepackage{enumitem}
\usepackage{comment}
\usepackage[pagebackref,breaklinks,colorlinks]{hyperref}

\usepackage[capitalize]{cleveref}
\crefname{section}{Sec.}{Secs.}
\Crefname{section}{Section}{Sections}
\Crefname{table}{Table}{Tables}
\crefname{table}{Tab.}{Tabs.}

\def\cvprPaperID{9622} 
\def\confName{CVPR}
\def\confYear{2023}

\begin{document}

\title{Attacking Object Detector Using A Universal Targeted Label-Switch Patch}

\author{Avishag Shapira, Ron Bitton, Dan Avraham, Alon Zolfi, Yuval Elovici, Asaf Shabtai\\
Department of Software and Information Systems Engineering\\
Ben-Gurion University of the Negev\\
}
\maketitle

\input{sections/00_abstract}
\input{sections/01_introduction}
\input{sections/03_background}
\input{sections/02_related_work_arxiv.tex}

\input{sections/04_method}

\input{sections/05_evaluation}

\input{sections/07_conclusion}

{\small
\bibliographystyle{ieee_fullname}
\bibliography{egbib}
}

\end{document}

%% file: sections/00_abstract.tex
\begin{abstract}
Adversarial attacks against deep learning-based object detectors (ODs) have been studied extensively in the past few years.
These attacks cause the model to make incorrect predictions 
by placing a patch containing an adversarial pattern on the target object or anywhere within the frame.
However, none of prior research proposed a misclassification attack on ODs, in which the patch is applied on the target object.
In this study, we propose a novel, universal, targeted, label-switch attack against the state-of-the-art object detector, YOLO.
In our attack, we use \textit{(i)} a tailored projection function to enable the placement of the adversarial patch on multiple target objects in the image (\eg, cars), each of which may be located a different distance away from the camera or have a different view angle relative to the camera, and \textit{(ii)} a unique loss function capable of changing the label of the attacked objects.
The proposed universal patch, which is trained in the digital domain, is transferable to the physical domain.
We performed an extensive evaluation using different types of object detectors, different video streams captured by different cameras, and various target classes, and evaluated different configurations of the adversarial patch in the physical domain.\\
 A short demo of our attack on a real car can be found at: \url{https://youtube.com/shorts/2A3CFQpgWGQ}.
 
\end{abstract}

%% file: sections/01_introduction.tex
\vspace{-15pt}
\section{\label{sec:intro}Introduction}

\begin{figure}[t]
    \captionsetup[subfigure]{labelformat=empty}
    \centering
    \begin{subfigure}{.48\linewidth}
        \centering
        \caption{Without patch}
        \includegraphics[width=\linewidth]{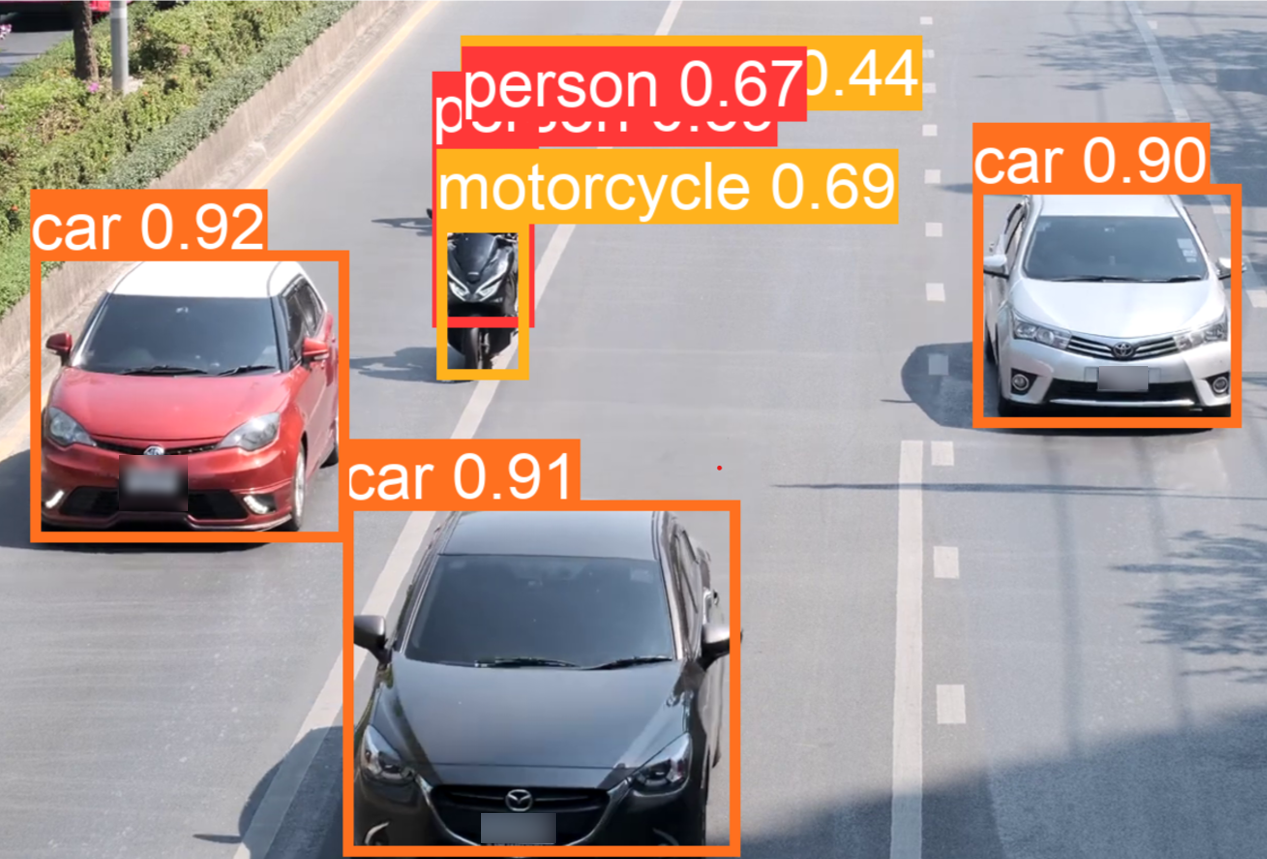}
    \end{subfigure}
    \hspace{0.05cm}
    \begin{subfigure}{.48\linewidth}
        \centering
        \caption{With patch}
        \includegraphics[width=\linewidth]{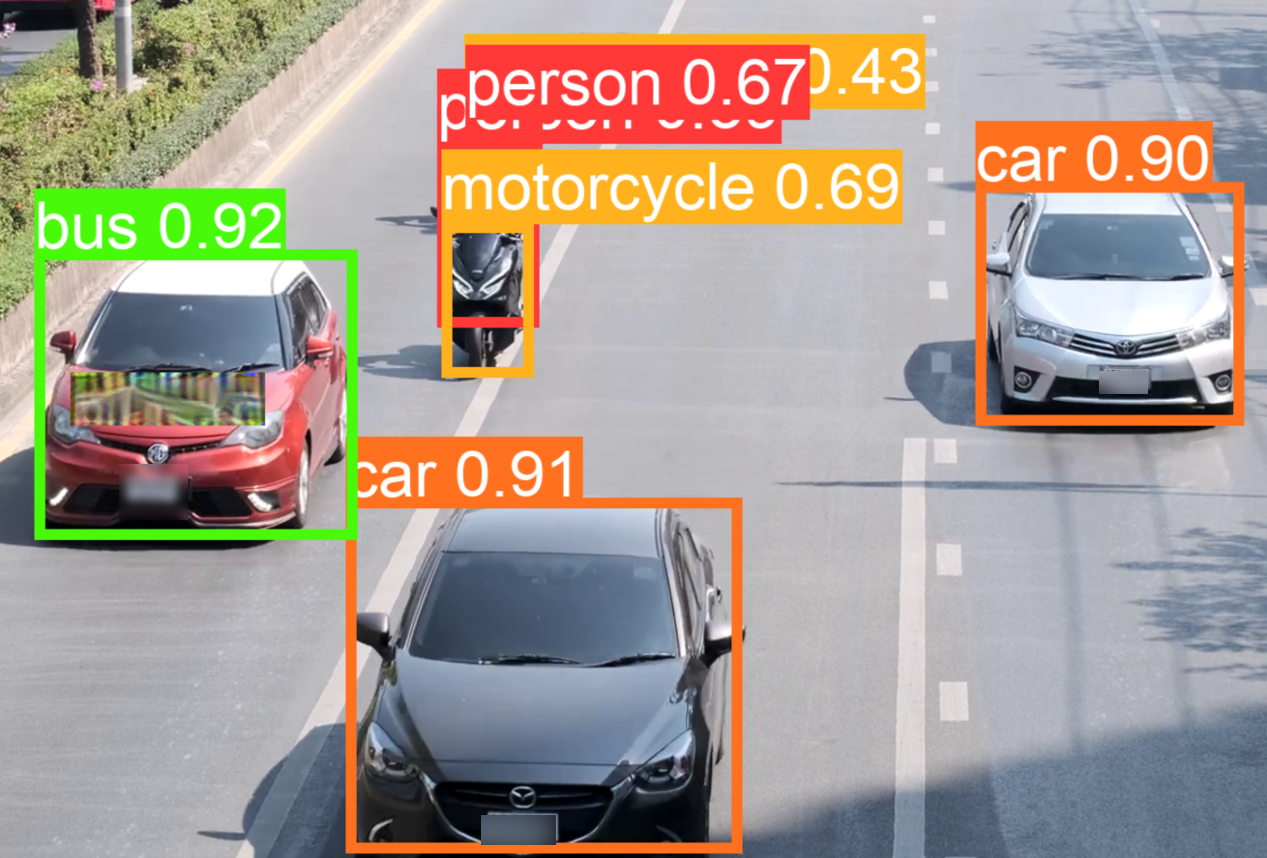}
    \end{subfigure}
    \hspace{0.05cm}
    \begin{subfigure}{.48\linewidth}
        \centering
        \includegraphics[width=\linewidth]{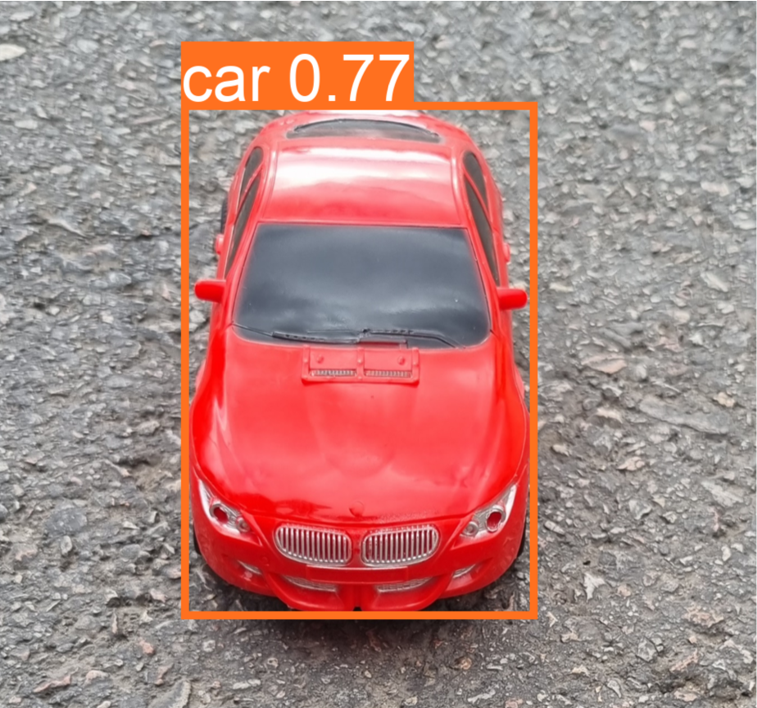}
    \end{subfigure}
    \hspace{0.05cm}
    \begin{subfigure}{.48\linewidth}
        \centering
        \includegraphics[width=\linewidth]{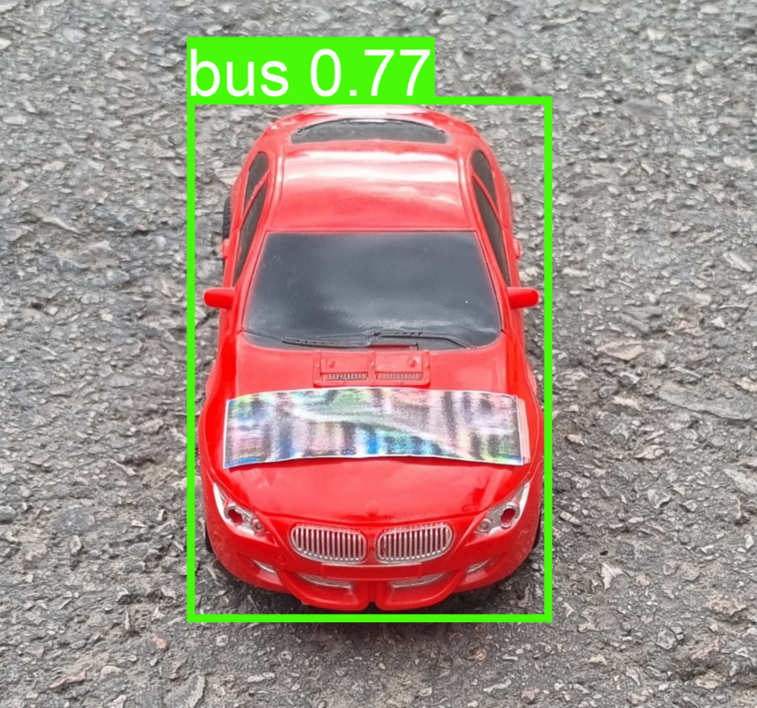}
    \end{subfigure}
    \caption{
    An illustration of our patch when applied in the digital domain (top) and physical domain (bottom).
    The clean (unattacked) image is on the left, and the attacked image, in which the patch is placed on a car, is on the right.
    The patch changes the car's classification from a car (in the orange bounding box) to a bus (in the green bounding box).
   }
    \label{fig:intro}
    \vspace{-10pt}
\end{figure}

Deep learning-based object detectors (ODs) such as YOLO~\cite{redmon2016you} and Faster-RCNN~\cite{ren2015faster} are widely used for various real-time applications~\cite{huang2018yolo,shafiee2017fast,laroca2018robust,lu2018vehicle,chen2019embedded,xu2018vehicle,maity2021faster} due to their ability to provide fast inference predictions with high accuracy.  
One use case in which ODs are playing an increasingly significant role is \emph{smart traffic systems}~\cite{yang2018vehicle}.
In this use case, cameras are used to monitor roads and intersections in order to optimize traffic flow, prioritize public transportation and emergency vehicles, and improve pedestrian safety in real time.
The video streams captured by the cameras are processed by a deep learning-based OD, and the objects identified are used by the smart traffic system to control and manage traffic lights.

Research performed over the last few years has shown that deep neural networks (DNNs) are vulnerable to adversarial machine learning attacks in which a patch that contains minor perturbations is crafted to cause misdetection~\cite{thys2019fooling} and misclassification~\cite{chen2018shapeshifter}.
Such patches can be placed on the object itself~\cite{thys2019fooling,xu2020adversarial}, anywhere within the frame~\cite{liu2018dpatch,zhao2020object}, or on the sensor (camera)~\cite{zolfi2021translucent}.

In this study, we propose a novel, universal, and targeted, label-switch attack against the state-of-the-art OD, YOLO~\cite{huang2018yolo}.
Specifically, we focus on the smart traffic system use case, where the goal of our attack is to cause the OD to classify a car on which the crafted patch is placed as a public transportation vehicle 
or an emergency vehicle.

We are the first to create a targeted \emph{misclassification} attack in which the patch is placed on the target object, making the attack both novel and realistic, unlike attacks that digitally place the patch within the frame in a non-realistic manner (\eg, DPatch~\cite{liu2018dpatch}) which are not transferable from the digital domain to the real world.
While other research which involved placing adversarial patches on the target objects was shown to be applicable in the physical domain, the attack aimed to \emph{hide} the objects~\cite{zhao2020object,eykholt2018robust}, whereas our proposed attack is aimed at misclassification which is very challenging to perform.

In our attack, we set out to achieve the following objectives: \textit{(i)} to cause the OD to classify a car containing the patch as another specific (predefined) type of object such as a bus, \textit{(ii)} to create a universal adversarial patch (UAP) that can be applied on any car, \textit{(iii)} to train the patch in the digital domain and successfully apply it in the physical domain, fooling the OD model with a high success rate, \textit{(vi)} to craft a patch which is robust to placement on different locations of the car relative to the camera's location, and \textit{(v)} to create an attack which is successful consistently over a sequence of frames for the same target car.

In order to achieve these goals, we designed our attack as follows.
First, we leveraged the fact that multiple cars typically exist in a scene, in order to create a (digital) patch that is robust to different locations of the car relative to the camera.
Specifically, we use a tailored projection function to place the adversarial patch on multiple target objects in the image (\ie, cars), each of which may be located in  a different distance away from the camera or have a different view angle relative to the camera.
This ensures both consistent misclassification of the same target object, \ie, car, throughout the sequence of frames in which the object appears and that the attack is effective for cars of any type/color.
Consequently, this results in a robust and universal patch in the digital domain which is transferable to and applicable in the physical domain.
Second, we designed a unique loss function that can change the label of the attacked objects, overcoming the challenge of objects that change their classification during the patch training process to classes other than car (the source class) or bus (the target class), which prevents the training process from converging.

We performed an evaluation in both the digital and physical domains, examining our attack on various types of ODs and testing the attack's transferability and effectiveness on various models simultaneously. 
In addition, we examined (1) the effectiveness of our attack when switching the detection of the attacked cars to different target classes, (2) the robustness of our patch when it is placed on cars that are in different locations in the frame, relative to the camera, or in different locations captured by a different camera, and (3) the performance of the patch when placing it on different parts of the car. 
The results show that our patch causes the misdetection success rate of over $90\%$ in both the physical and digital domains. 
Our attack can also craft a patch that successfully attacks multiple models simultaneously and is effective when placed on different locations of the car.

We summarize the contributions of our work as follows:
    (1) We are the first to propose a novel, universal, and targeted, label-switch attack against a state-of-the-art object detector. 
    (2) We propose a tailored and practical projection function which enables the placement of the adversarial patch on multiple target objects, \ie, cars, in the image each of which may be located a different distance away from the camera or have a different view angle relative to the camera.
    (3)  We present a unique loss function capable of changing the label of the attacked objects, overcoming the challenge of objects changing their label to unwanted classes during the patch training process. 
    (4) We demonstrate the patch's robustness in both the digital and physical domains as well as its transferability to different models.
   

%% file: sections/03_background.tex
\section{\label{sec:background}Background}

\noindent\textbf{\label{subsec:yolo}YOLO Object Detector} ~\cite{redmon2016you,redmon2017yolo9000,redmon2018yolov3}, the state-of-the-art one-stage OD is the target of our proposed attack. 
The architecture of many subsequent ODs (\eg, YOLOv4~\cite{bochkovskiy2020yolov4} and YOLOv5~\cite{yolov5}) derived from YOLO's architecture design.
YOLO's architecture consists of two parts: a convolution-based backbone used for feature extraction, which is followed by \emph{multi-scale} grid-based detection heads used to predict bounding boxes and their associated labels.
A 3D tensor is predicted by the last layer of each of the detection heads; this 3D tensor encodes the following: bounding box (the coordinate offsets from the anchor box); the objectness score (the detector's confidence that the bounding box contains an object ($Pr(Object)$)); and the class scores (the detector's confidence that the bounding box contains an object of a specific class ($Pr(Class_i \vert Object)$)).
For a given image size, a fixed number of candidate predictions are produced by YOLO, and a predefined threshold $T_{\text{conf}}$ is utilized to filter these predictions based on the predictions' objectness score and class score (the result of multiplying these values needs to be higher than $T_{\text{conf}}$).

Finally, since many candidate predictions may overlap and predict the same object, the \textbf{Non-Maximum Suppression (NMS)} algorithm is applied to remove redundant predictions.
The intersection over union (IoU) value, calculated as the ratio of the overlap area divided by the union area for each pair of candidates, is used by the NMS algorithm to identify overlapping candidates. 
Two candidates that have the same target class are considered to be overlapping candidates when their IoU value is greater than a predefined threshold ($T_{\text{IoU}}$).

%% file: sections/02_related_work_arxiv.tex
\section{\label{sec:related}Related Work}

Previous studies presenting adversarial attacks on ODs can be categorized by the goal of the attack.
Some attacks focused on hiding specific objects, such as stop signs~\cite{zolfi2021translucent}, people~\cite{xu2020adversarial,thys2019fooling} or any object~\cite{zhao2020object,liu2018dpatch} from the OD, while others aimed at causing the system to misclassify an object as another object (any object or a specific object, \ie, target object)~\cite{liu2018dpatch,brown2017adversarial,sitawarin2018rogue,chen2018shapeshifter,eykholt2018robust}.
In this paper, we propose a misclassification attack on an object detection system.

Prior work can also be categorized by the location in which the patch is placed.
In some attacks the patch was placed in the frame, \eg, in the corner of the frame~\cite{liu2018dpatch} or randomly in a single location~\cite{liu2018dpatch,brown2017adversarial} or multiple locations~\cite{zhao2020object} in the frame.
In the study performed by Zolfi \etal~\cite{zolfi2021translucent} the patch was placed on the sensor itself, \ie, the camera's lens, while Sitawarin \etal~\cite{sitawarin2018rogue} and Chen \etal~\cite{chen2018shapeshifter} produced a new object on which a perturbation is applied.
In this research, we are interested in applying the patch on a specific object, so that it can be applied in the real world, unlike the case where the patch is applied in a location on the frame (which is not transferable to the real world).

Table~\ref{tab:related} which compares the attack methods mentioned above based on the following properties: \textit{i)} attack goal -- misclassification, hiding objects, or adding phantom objects; \textit{ii)} patch placement -- anywhere in the frame, on the target object(s), on the sensor, or on an object created for this purpose; \textit{iii)} machine learning (ML) task -- object detection or image classification; \textit{iv)} consider changing view point relative to the camera and movement of the object in a sequence of consecutive frames; \textit{v)} attacking multiple objects in the frame simultaneously; \textit{vi)} assuming one main object in the attacked frame during the patch training process; \textit{vii)} testing the attack in the physical domain; and \textit{viii)} testing the transferability of the attack across multiple models.

As can be seen from the table, none of the studies proposed a misclassification attack on ODs in which the patch is applied on the object itself and maintains its robustness to various locations of the attacked object relative to the camera in both the digital and physical domains.
This is due to the challenges (resented in Section~\ref{sec:method}) in performing this attack.

\input{sections/02_relatedworksTable.tex}

%% file: sections/02_relatedworksTable.tex
\begin{table*}[t!]
\scriptsize
\centering
\begin{tabular}{|p{1.55cm}|p{1cm}|p{1.15cm}|p{0.60cm}|p{1.95cm}|p{1.3cm}|p{1.5cm}|p{1.1cm}|p{1.1cm}|p{2.45cm}|}
\hline
& Attack goal$^{(1)}$ & Placement$^{(2)}$ & ML task$^{(3)}$ & Consider changing view point (sequence of frames) & Attack multiple objects & Assume one main object in the frame & Test in physical domain & Test transferability & Model \\ \hline\hline

DPatch~\cite{liu2018dpatch} & M / H / P  & F 
& OD & $\times$ & \checkmark & $\times$ & $\times$ & \checkmark & YOLO, Faster R-CNN\\ \hline

Adv. Patch~\cite{brown2017adversarial} & M & F 
& C & $\times$ & $\times$ & \checkmark & \checkmark & \checkmark & Incep.V3, ResNet50, Xception, VGG16/19\\ \hline

Obj. Hider~\cite{zhao2020object} & H & F 
& OD & $\times$ & \checkmark & $\times$ & $\times$ & \checkmark & YOLO, Faster R-CNN\\ \hline

Translucent Patch~\cite{zolfi2021translucent} & H & S & OD & $\times$ & \checkmark & $\times$ & \checkmark & \checkmark & YOLOv5, YOLOv2, Faster R-CNN\\ \hline

Rogue Signs~\cite{sitawarin2018rogue} & M & NO 
& C & Perspective transformation & $\times$ & \checkmark & \checkmark & $\times$ & Custom GTSRB, Multi-Scale CNN\\ \hline

Shape Shifter~\cite{chen2018shapeshifter} & M & NO 
& OD & Trained using stop signs from different distances/angels & $\times$ & \checkmark & \checkmark & \checkmark & Faster R-CNN\\ \hline

Physical-World Attacks~\cite{eykholt2018robust} & M & O & C & Trained using stop signs from different distances/angels & $\times$ & \checkmark & \checkmark & \checkmark & LISA-CNN, GTSRB-CN \\ \hline

Attack Person Detection~\cite{thys2019fooling} & H & O & OD & $\times$ & $\times$ & \checkmark & \checkmark & \checkmark & YOLOv2, Faster R-CNN \\ \hline

Adv. T-Shirt~\cite{xu2020adversarial} & H & O & OD & Trained on moving people & $\times$ & \checkmark & \checkmark & \checkmark & YOLOv2, Faster R-CNN \\  \hline

\hline \hline
Our method & M & O & OD & \checkmark & \checkmark & $\times$ & \checkmark & \checkmark  & YOLO, Faster R-CNN \\ \hline
\end{tabular}
\caption{Comparison of attack methods.}
\label{tab:related}
\scriptsize{$^{(1)}$M -- Misclassification; H -- Hiding objects; P -- Adding phantom objects. \\}
\scriptsize{$^{(2)}$F -- Anywhere in the frame; O -- On the target object(s); S -- On the sensor (\ie, camera); NO -- Creating new object to fool the model.\\}
\scriptsize{$^{(3)}$OD -- Object detection; C -- Classification.\\}
\end{table*}

%% file: sections/04_method.tex
\section{\label{sec:method}Method}

\begin{figure*}[t!]
\centering
    \includegraphics[width=0.85\linewidth]{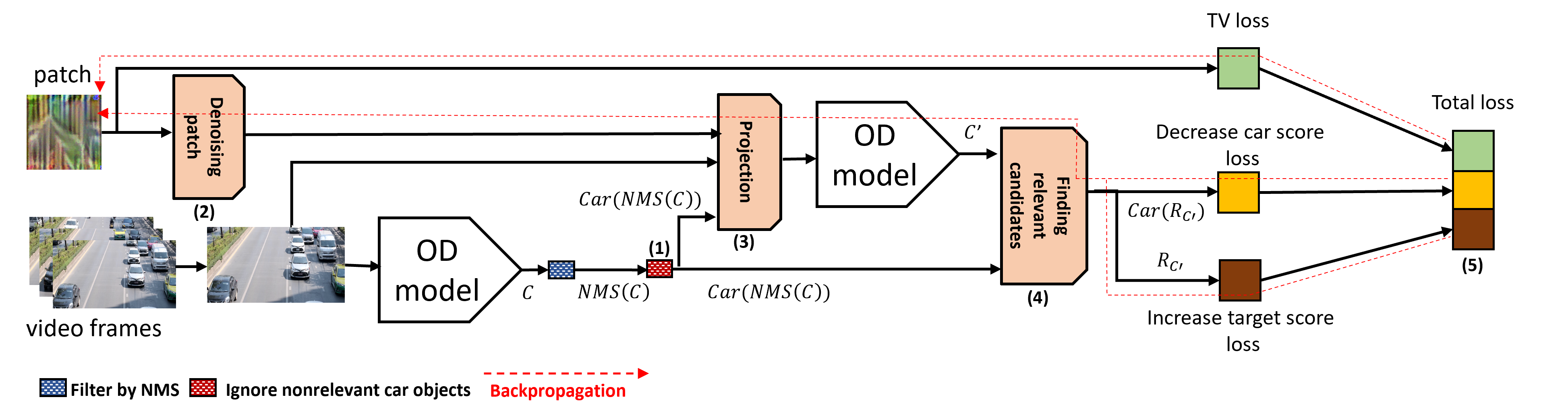}
    \caption{Overview of our method's pipeline.}
    \label{fig:pipeline}
\end{figure*}

\noindent\textbf{Goal.} We aim to produce a \emph{universal}, targeted, physical patch that changes the detection of \emph{any} car containing the patch (the real class) to another target class (\eg, bus, truck, or emergency vehicle).
We also want to train the patch in the digital domain and to be able to apply it successfully (\ie, the attack will fool the OD model with a high success rate) in the physical domain. 
We do this in order to avoid the complexity involving the identification of the patch's location and the optimization of the patch when \emph{performing the training process in the physical domain}.
Finally, since we are focusing on the use case of a smart traffic management system, in which the camera is static, and the cars' movements cause the distance and view angle from the camera to change, we aim to create a patch that is robust to different locations of the car relative to the camera.
In addition, the attack should be successful consistently in a sequence of frames for the same target car containing the patch.\\
\noindent\textbf{\label{subsec:challenges}Challenges.}
Applying a targeted patch on a car that fools the OD model in videos from a surveillance camera perspective presents the following challenges:
\begin{enumerate}[label=(\roman*),noitemsep,topsep=0pt]
    \item The size and angle of the parts of the vehicle, as well as the location of those parts within the vehicle's bounding box, changes depending on the location of the vehicle in an image (which is relative to the camera's location.
    Therefore, we need to adjust our projection of the patch on each car with respect to the car's location and bounding box area.
    
    \item During the training process, the patch might change the candidates' detection to an unwanted class that differs from car (source class) or bus (target class), or even cause the candidates to be undetected by the OD model. 
    Therefore, it is not sufficient to consider just the candidates that were detected as a car; thus, an approach that identifies and makes use of all of the relevant candidates during the training process is required.
    
    \item Since we train our patch on a video stream which includes moving vehicles, there are many cases where part of a car appears to be ``cut out" of the frame. 
    Training the patch when including the cars that partially appear in the frame harms the training process and reduces the effectiveness of the derived patch. 
    Therefore, we must determine how to treat these cars during the training phase, for example, by ignoring cars that are mostly out of the frame.

    \item To make our attack stronger (which will place less restrictions and requirements on the attacker), we aim to create a universal patch that will be able to cause the misclassification of any car and is transferable to different locations (\ie, cameras) and models.
\end{enumerate}
\noindent\textbf{Threat Model.} In our case, we assume that the attacker has access to a video stream captured by the camera he/she wants to fool (a few minutes of recorded video should be sufficient) or alternatively, to a video stream from a camera from another location with a similar position and view angle relative to the road (and the cars on the road).
We consider this threat model practical given the large number of real-time video stream from real road cameras available on the Internet and video platforms (either real-time videos or offline recorded ones).
In addition, we assume that the attacker knows that the OD model used by the smart traffic system is one of several commonly used models, but he/she does not know the exact model. \\
\noindent\textbf{Attack Overview.} During the training phase, we apply our patch on multiple cars in the video frames from a surveillance camera and update our patch using a gradient-based optimization method to decrease the loss function.
A single iteration of our attack (\ie, patch training process) is illustrated in Figure~\ref{fig:pipeline}.
In each iteration, we perform the following steps:
(1) We identify the car objects in the clean (unattacked) frame on which we want to apply our patch, ignoring car objects that are too small or ``cut out" of the frame. 
(2) We add random noise to the patch, also known as patch denoising~\cite{thys2019fooling}, to make our patch more robust in the physical domain.
(3) We apply (project) the patch on each of the selected car objects, while considering the shape of the car's hood and the car's relative location in the frame. 
(4) Before the NMS stage, we identify the relevant candidates in the OD output in the attacked frame, \ie, the frame with the patch applied on the car objects; this is performed based on their IoU value with the selected car objects in the clean frame.
This is an important step, since the patch might change the classification of the car objects, and therefore we cannot rely on the output of the OD on the attacked frame for the identification of the car objects. 
(5) We compute the total loss function using these candidates and the patch pixels values, and update the patch accordingly.

\input{sections/04_method_projection.tex}

\input{sections/04_method_candSelection.tex}

\input{sections/04_method_loss.tex}

%% file: sections/04_method_projection.tex
\subsection{\label{subsec:project_patch}Patch Projection}

In order to successfully train and optimize the patch in the digital domain, we need to adjust the relative location of our patch with respect to the location of the car, \ie, considering the $X$ and $Y$ axes. 
We must also determine the patch shape and size with respect to the cars we wish to apply our patch on.
In addition, to craft a patch that works efficiently in the physical domain, we need to add projection elements that simulate real-world constraints.

\noindent\textbf{\label{subsec:patch_denoising}Patch Denoising.}
Similar to Thys \etal~\cite{thys2019fooling}, in order to make our patch more robust and transferable to the physical domain, we add random Gaussian noise to the patch and change the patch's brightness and contrast randomly.

\begin{figure}[h]
    \centering
    \includegraphics[width=0.99\linewidth]{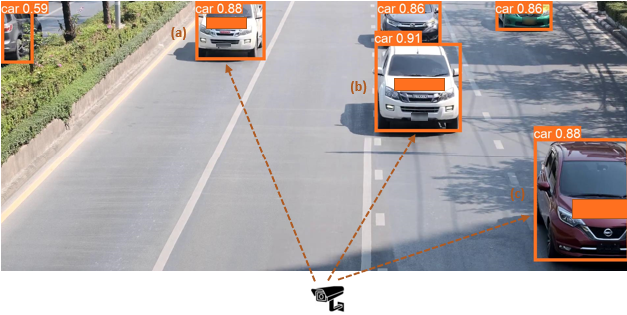}
    \caption{Relative hood location for cars at different angles and distances captured by the surveillance camera.}
    \label{fig:angles}
\end{figure}

\begin{figure}[h]
    \centering
    \includegraphics[width=0.40\linewidth]{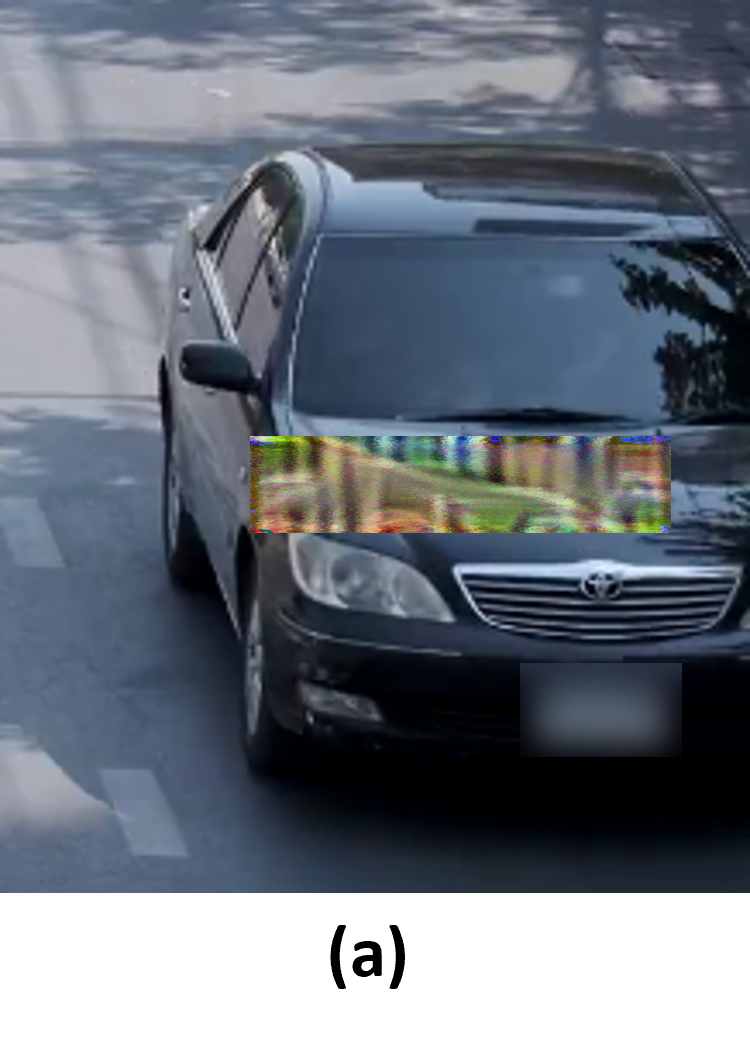}
    \includegraphics[width=0.40\linewidth]{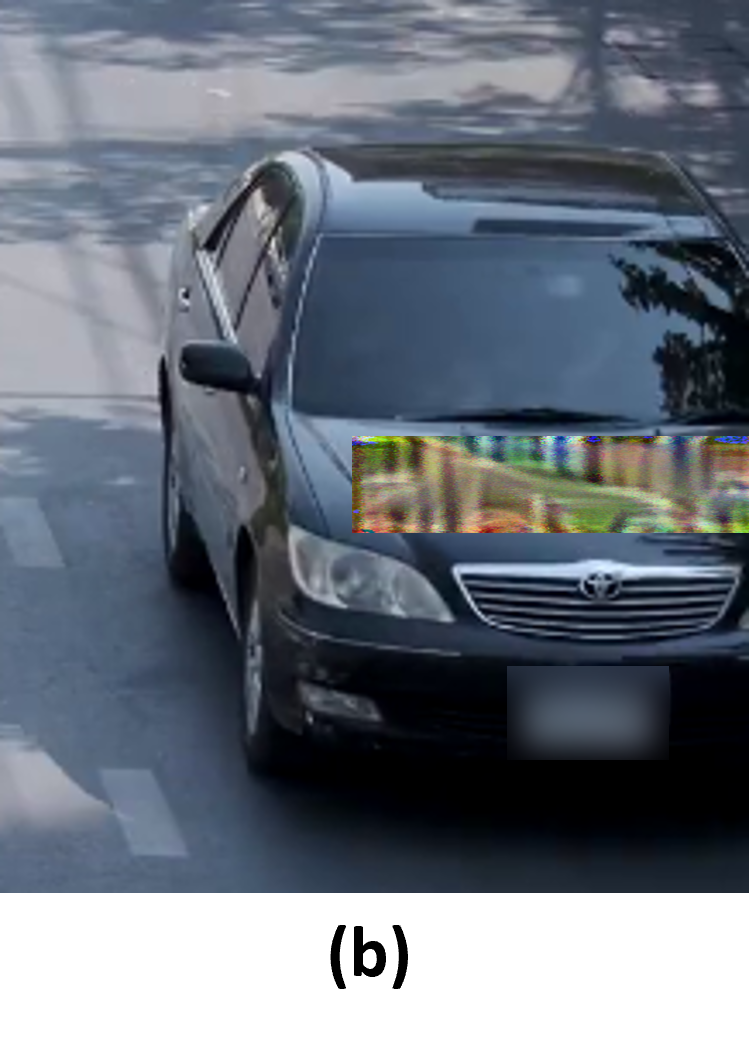}
    \caption{Projecting our patch using the previous approach (a) and our projection function (b).}
    \label{fig:x_proj}
\end{figure}

\noindent\textbf{\label{subsec:patch_shape}Determine the patch shape and size.}
We opted to place our patch on the car's hood, because it is a part of the vehicle that is relatively visible from a distance and from different camera angles and perspectives. 
Since most cars' hood are rectangular, the patch we project should also be rectangular. 
In our preliminary experiments, we observed that a trained square patch (pixel-wise) which is stretched to a rectangle using an affine transformation~\cite{thys2019fooling} leads to better results than training a rectangular patch.
For each car in the frame, we determine the size of our patch based on the relative size of the hood in an average car object in the video. \\
\noindent\textbf{\label{subsec:ignore}Cars ``cut out" of the frame.}
Some of the car objects identified by the OD may be partially ``cut out" of the frame, \eg, when the car is entering or exiting the frame.
In such cases, the bounding boxes of these car objects are ``too rectangular," \ie, the ratio between the object's height and the object's width is very low or high.
Projecting the patch on these partial car objects is challenging and may lead to a patch that does not successfully transfer to the physical domain.
Therefore, we opt to ignore cars for which the ratio between the object's height and width is very low or high (above a predefined threshold). 
In addition, we ignore cars whose bounding box size is very small; these are cases in which either the OD detects a small portion of the car or the car is very far away from the camera. 
In both cases, we do not want to apply our patch on the car, since either the patch cannot be applied on the hood in a realistic manner or the patch cannot be clearly seen by the camera (since the car is too far away).\\
\noindent\textbf{\label{subsec:project_x}Projecting on the $X$-axis.}
The location of the car's hood with respect to the car's bounding box changes according to the location of the car within the frame. 
When the car is located more towards the center of the frame (see car (b) in Figure~\ref{fig:angles}), or when the car is further away from the camera (car (a) in Figure~\ref{fig:angles}), the car's hood will be located in the center of the car's bounding box. 
However, when the car is located near the edge of the frame and closer to the camera, the camera captures the sides of the car, and therefore, the location of the car's hood changes relative to the center of the car's bounding box (as can be seen with car (c) in Figure~\ref{fig:angles}). 
The same can be observed for a car that is partially ``cut out" from the frame (but still mainly inside it).
In this case, since the OD model detects just part of the car (the part that is inside frame), the car's hood will not be located in the center of the car's bounding box (as can be seen with car (c) in Figure~\ref{fig:angles}). \\
Therefore, using the previously used projection approach~\cite{thys2019fooling} to place the patch will result in an unrealistic projection (\ie, cannot be placed in the real-world), since the patch will be placed in the center of the bounding box instead of being centered on the car's hood.
Figure~\ref{fig:x_proj} illustrates this case: on the left side (a) we present the projection of the patch utilizing the previously used approach, and on the right side (b) we present the projection of the patch utilizing our tailored projection function.\\
In our projection function, we change the patch's position on the $X$-axis as follows:
\begin{equation}
 x_{proj} = x + (\alpha \cdot (y \cdot\max(x, 1-x))^2\cdot(x-0.5))
\end{equation}
\noindent where $x$ and $y$ are the values of the center of the car's bounding box on the $X$- and $Y$-axes respectively (normalized to values between zero and one).\\ 
$x_{proj}$ is the value of the center of our projected patch on the $X$-axis, and $\alpha$ is a weighting factor, which controls the rate of the patch's movement. 
This value can change for different surveillance cameras (\ie, scenes) that capture a different range of angles. \\
$y \cdot \max(x, 1-x)$ results in a higher value for cars that are close to edge of the frame and close to the camera. 
We compute the squared power of this value in order to reduce the rate at which the patch is ``pushed" away from the center of the bounding boxes of cars' that are close to the center of the frame or far from the camera. 
Since a value of 0.5 represents the bounding box's center, we multiply the value above by $(x-0.5)$, in order to determine which direction the patch should be pushed towards. 

\noindent\textbf{\label{subsec:project_y}Projecting on $Y$-axis.}
When a car is closer to the camera, the relative location of its hood within the detected bounding box is much lower compared to the relative location of the hood of a car that is further from the camera (for example, car (b) and car (c) in Figure~\ref{fig:angles}). 
In addition, the patch's relative position changes faster for car objects that are closer to the camera, because their angle relative to the camera changes faster.
Therefore, we make two adjustments to the projection on the $Y$-axis:
(1) The farther the car is from the camera (\ie, small $y$ value), we ``raise" the relative position of the patch (by decreasing its $y$ value), and conversely, the closer the car is to the camera, we ``lower" the relative position of the patch (by increasing its $y$ value).
(2) For cars that are close to the camera, we ``lower" the patch position at a faster rate than the ``rate at which it rises for cars that are farther away.
Therefore, we use the $y$ value of the bounding box as a weighting factor that determines how large the shift of the patch is.
The relative position of the patch on the $Y$-axis is determined in the following way:
\begin{equation}
  y_{proj}= y +  \beta \cdot y^2 * (y - 0.5)
\end{equation}
\noindent where $y$ is the value of the center of the car's bounding box on the $Y$-axis, $y_{proj}$ is the value of the center of our projected patch on the $X$-axis, and $\beta$ is a weighting factor, which controls the rate of the patch's movement. 
This value can change for different surveillance cameras that capture vehicles at different distances.\\
$y^2$ results in a higher value for cars that are closer to the camera. 
Since the value of 0.5 represents the bounding box's center, we multiply the value above by $(y-0.5)$, in order to determine which direction the patch should be pushed towards.

%% file: sections/04_method_candSelection.tex
\subsection{\label{subsec:relevant_can}Finding Relevant Candidates}

As mentioned, during the patch optimization process, applying the patch on an object (\ie, car) might change the object's classification. 
If during the training phase only the candidates that the OD detects as cars are considered (as was done in~\cite{thys2019fooling}), many of the relevant candidates are ignored, which results in a less effective patch that is not successful in changing the classification of a car. 
Therefore, instead of determining the relevant candidates based on their target class provided by the OD, we identity relevant candidates in the attacked frame based on their IoU value with car objects in the clean frame.

In addition, for each predicted car object in the clean frame, there can be several overlapping candidates that pass the confidence score filtering phase (described in Section~\ref{subsec:yolo}) during the inference stage, which are then reduced by the NMS component to a single final candidate. 
Since the NMS stage only reduces overlapping candidates that have the same target class, we want our patch to change the detection of all of these candidates to the desired target class (in our case, bus).
We would also like to avoid the case in which the patch changes the detection for just some of these candidates, which may lead to a situation of ``double detection" (\ie, the car will be detected by the OD as both car and bus).
Therefore, instead of comparing the IoU between the final candidates in the clean frame and the final candidates in the attacked frame, we compare the IoU between the final candidates in the clean frame and the candidates \emph{before} applying the NMS component on the attacked frame.

More formally, let: (1) $C$ and $C'$ be the candidates in the clean and attacked frames respectively before applying the NMS component (regardless their target class); (2) $NMS(C_G)$ be the final candidates (after the NMS component) for a group of candidates $C_G$; 
(3)  $c_{\text{score}\,i}$ be the class score of a candidate $c \in C_G$ for a target class $i$; (4) ${Car(C_G)=\{{\max\{c_{\text{score}\,i}\}_{i=0}^{N_c} = c_{\text{score}\,car} \vert c \in C_G}}\}$ be the candidates that the OD detected as car; and (5) $T_{IoU}$ be a predefined threshold that defines the minimum IoU value for which candidates are treated as overlapping candidates (and eventually reduced by the NMS component to a single candidate). 

\noindent The relevant candidates in the attacked frame for a single final object $c$ in the clean image are defined as:
\begin{equation}
    R_{C'}(c)={\{\text{IoU}(c, c') >= T_{\text{IoU}} \vert c' \in C'\}}
\end{equation}
\noindent To find the relevant candidates in the attacked frame for all of the final car objects in the clean image, we use the following expression:
\begin{equation}
 R_{C'}= \cup_{c \in Car(NMS(\mathcal{C}))}{( R_{C'}(c))}
\end{equation}

%% file: sections/04_method_loss.tex
\subsection{\label{subsec:optimization_process}Optimization Process}

Our attack aims to produce a patch that switches the detection of a car object that the patch is applied on (referred to as $c'$) to a target class $t$. 
In order to achieve this goal, the patch should ensure that the target class $t$ receives the maximal class score, \ie, ${\max\{{c'}_{\text{score}\,i}\}_{i=0}^{N_c} = {c'}_{\text{score}\,t}}$. 

To ensure that our attack fulfills this goal efficiently, it needs to perform the two following actions at the same time: \textit{i)} decrease ${c'}_{\text{score}\,car}$ and \textit{ii)} increase ${c'}_{\text{score}\,t}$.
To accomplish this, we use binary cross-entropy loss $\mathcal{L}_{BCE}$ to represent the distance between the current candidates' score and the desired score.
Specifically, we define a ``target" candidate $c$ which represents a binary vector where ${c_{\textit{score i}}=0}$ for all $i\neq t$ and ${c_{\textit{score t}}=1}$.\\ 
Note, that in order to find all the relevant candidates during the training phase, and avoid cases where relevant candidates ``disappear," (\ie, these candidates are filtered before the NMS component), we need to ensure that  ${c'}_{\text{score}\,car}$ does not decrease ``too fast" before the ${c'}_{\text{score}\,t}$ increases. 
Therefore, we prioritize increasing the ${c'}_{\text{score}\,t}$ value over decreasing the ${c'}_{\text{score}\,car}$ value during the training phase (using a weighing factor).\\
\noindent\textbf{\label{subsec:decrasing_car}Decrease-car-score loss.}
In this component, we want to decrease ${c'}_{\text{score}\,car}$ for candidates that the OD model detects as car ($Car( R_{C'})$), since these candidates probably have a high  ${c'}_{\text{score}\,car}$ value.
We average the loss value of all  {$c' \in Car(R_{C'})$} candidates:
\begin{multline}
    \ell_{\text{decrease car}} = \frac{1}{\vert Car(R_{C'})\vert}\cdot \\
    \sum\limits_{c' \in Car(R_{C'}))} 
    \mathcal{L}_{\text{BCE}}({c'}_{\text{score}\,car},{c}_{\text{score}\,car})
\end{multline}

\noindent\textbf{\label{subsec:incrasing_target}Increase-target-score loss.}
In this component, we want to increase ${c'}_{\text{score}\,t}$ for all of the relevant candidates found in the attacked frame, regardless of their target class, in order to increase the probability that the OD model detects the candidates as the target class $t$.
We average the loss value of all  {$c' \in R_{C'}$} candidates:
\begin{equation}
    \ell_{\text{increase target}} = \frac{1}{\vert R_{C'}\vert}\cdot
    \sum\limits_{c' \in R_{C'}} 
    \mathcal{L}_{\text{BCE}}({c'}_{\text{score}\,t},{c}_{\text{score}\,t})
\end{equation}

\noindent\textbf{\label{subsec:decrasing_TV}Total variation loss.}
Since we want to use our patch in the physical world, in our loss function we consider the total variation (TV) factor~\cite{shao2019objects365}, which ensures that our attack crafts a patch with smooth color transitions between neighboring pixels:
\begin{equation}
    \ell_{\text{TV}} =
    \sum\limits_{i,k} 
    \sqrt{(p_{i,k} - p_{i+1,k})^2 + (p_{i,k} - p_{i,k+1})^2}
\end{equation}
$p_{i,k}$ represents the value of pixel in $(i,k)$ location in our patch.

\noindent\textbf{\label{subsec:loss}Final loss function.}
Finally, the optimization problem we solve aims to find a patch $p$ that minimizes the following expression:
\begin{equation}
    \min_{p}{[\lambda_1 \cdot \ell_{\text{decrease car}} + (1-\lambda_1) \cdot \ell_{\text{increase target}} + \lambda_2\ell_{\text{TV}}]}
\end{equation}
\noindent where $\lambda_1$ is a weighting factor that balances between increasing candidates' car score and decreasing candidates' target score, and $\lambda_2$ is a weighting factor that controls the smoothness of the patch crafted by our attack.\\

\noindent\textbf{\label{subsec:ensemble}Ensemble training.}
To improve the transferability of our attack to different OD models, we perform ensemble training using $K$ models, where in each iteration the average loss value is computed for all ${k\in K}$ to backpropagate and update the patch pixels.

%% file: sections/05_evaluation.tex
\section{\label{sec:eval}Evaluation}
We evaluate the effectiveness of our patch in both the digital and physical domains. 

\subsection{Evaluation Setup}
\noindent\textbf{Models.}
We evaluated our attack using the following versions of YOLO: YOLOv3~\cite{redmon2018yolov3}, YOLOv4~\cite{bochkovskiy2020yolov4}, and YOLOv5~\cite{yolov5}, and Faster R-CNN~\cite{ren2015faster}.

\noindent\textbf{Data.}
For the training phase and the digital evaluation, we used three video clips taken from different vehicle surveillance cameras (downloaded from a large video content provider on the Internet): (1) an 8-minute video clip (denoted by v1), (2) a 6-minute video clip (v2), and (3) a 14-second video clip (v3).
We divide v1 and v2 into 2,000 frames each, where 1,500 frames are used in the training phase, and the remaining 500 frames are used in the digital evaluation process.
V3 is only used in the evaluation process.

\noindent\textbf{Implementation details.}
Our patch size was set at $300X300$ pixels.
For patch denoising we changed the patch's brightness within the range of $0.8$-$1.2$, the patch's contrast change range was set at $-0.1$-$0.1$, and the Gaussian noise factor was defined as 0.1.\\
For v1 and v3, $\alpha=0.2$ and $\beta=1$ to achieve optimal projection, while for v2,  $\alpha=0.1$, and $\beta=0.8$.\\
In our loss function, $\lambda_1=0.2$ and $\lambda_2=3$ values.
For the target models, we used the small sized YOLOv5 version (YOLOv5s), as well as YOLOv3 and YOLOv4.
For the ensemble learning, different combinations of these three YOLO models were used.
We also examined the transferability of our patch on a Faster-RCNN model.\\
In the inference stage, we set $T_{conf}=0.25$ (a minimum score threshold for considering a candidate as an object) and $T_{IoU}=0.45$ (a minimum threshold to define two candidates as overlapping candidates), since they are the default values commonly used for these models in the inference phase.\\
The \emph{bus} class is our attack's target class.\\
Unless mentioned otherwise, the patches were trained and evaluated on a YOLOv5s model with v1 frames.

\noindent\textbf{Evaluation metrics.}
To evaluate our patch's performance, we used two metrics: \textit{i)}  $C_t\%$ - the percentage of car objects on which a patch is placed whose label was switched to class $t$. 
\textit{ii)} Double detection - the percentage of car objects on which a patch is placed that the OD model detects both as car and bus (\ie, the attack target class) (as explained in Section~\ref{subsec:relevant_can}).

\subsection{Results: Digital Experiments}
\noindent\textbf{Effectiveness of our patch.}
We projected our patch on the hoods of 1,084 car objects (detected by YOLOv5s in the 500 test frames of v1), 509 car objects (detected by YOLOv5s in the 500 test frames of v2), and 217 car objects (detected by YOLOv5s in the 150 frames of v3).
We compare the results of our patch with a baseline attack -- a patch created with randomly colored pixels sampled uniformly. 
As seen in Table~\ref{table:base} which presents the results, the crafted patches caused a switch in the target class of 88\% - 94.6\% of the detected car objects from that of car to bus, while the success rate of the random patch was 8.5\% on average.
The results also demonstrate the robustness of the patch across different scenes and cameras (\ie, when training the patch on the video and testing on a video from another camera). 

\begin{table}[h]
\centering
\scalebox{0.7}{
\begin{tabular}{lccc}
\hline
       & \multicolumn{3}{c}{Tested video clip}             \\
       & v1          & v2          & v3        \\
       & \multicolumn{3}{c}{C\_bus\% / double detection\%} \\ \hline \hline
Random & 4.2 / 7.1       & 13.3 / 14.7     & 6.9 / 8.3     \\
v1 & 94.6 / 12.2     & 89.8 / 10.1     & 93.5 / 5.5    \\
v2 & 89.4 / 9.1      & 90.1 / 13.8     & 88.0 / 7.9     
\end{tabular}
}
\caption{Attack success rates (digital domain) when training and testing the patch on different video clips.}
\label{table:base}
\end{table}

\noindent\textbf{Different models.}
We crafted patches for three different versions of YOLO (YOLOv3, YOLOv4, and YOLOv5s) to demonstrate the universality of our attack. 
The results presented in Table~\ref{table:trans_table} show that the attack is effective on all three models.

\begin{table}[h]
\centering
\scalebox{0.7}{
\begin{tabular}{lcccl}
\hline
       & \multicolumn{3}{c}{Victim Models}   &    \\
       & YOLOv3   & YOLOv4  & YOLOv5s    & Faster-RCNN \\
       & \multicolumn{4}{c}{C\_bus\% / double detection\%}                              \\ \hline \hline
Random  &  3.1/ 40.6 &  3.2 / 0         &  4.2 / 7.1           &  5.5 / 14.5 \\
YOLO3  & \textbf{90.5 / 14.4} & 42.5 / 27.6        & 40.3 / 61.3          & 40.3 / 58.3 \\
YOLOv4 & 23.1 / 60.0            & \textbf{88.0 / 12.3} & 19.7 / 41.1            & 17.9 / 45.2 \\
YOLOv5s & 9.7 / 31.2           & 10.4 / 43.0          & \textbf{94.6 / 12.2} & 47.8 / 46.7
\end{tabular}
}
\caption{The performance of patches trained for different versions of YOLO.}
\label{table:trans_table}
\end{table}

\noindent\textbf{Transferability and Ensemble Learning.}
As can be seen in Table~\ref{table:trans_table}, the proposed attack is partially transferable between different models.
Therefore, to improve the transferability between models we opted to use an ensemble learning technique, \ie, train the patch on multiple models simultaneously. 
The results of the ensemble approach, which are presented in Table~\ref{tab:ens}, 
demonstrate the effectiveness of patches created using the ensemble technique. These results show that an attacker trying to perform the attack does not need to know the exact type/version of the model; in order to  perform a successful attack, one patch trained on an ensemble of models can be effective.\\

\begin{table}[h]
\centering
\scalebox{0.7}{
\begin{tabular}{lcccc}
\hline
               & \multicolumn{3}{c}{Victim Models}                                  & \multicolumn{1}{l}{}            \\
               & YOLOv3               & YOLOv4               & YOLOv5s              & \multicolumn{1}{l}{Faster-RCNN} \\
               & \multicolumn{4}{c}{C\_bus\% / double detection\%}                                                    \\ \hline \hline
YOLOv5s        & 9.7 / 31.2           & 10.4 / 43.0            & \textbf{94.6 / 12.2} & 47.8 / 46.7                     \\
$\text{Ens}_1$ & 29.7 / 35.3          & \textbf{88 / 23.1}   & \textbf{85.6 / 24.4} & 49.0 / 56.1                       \\
$\text{Ens}_2$ & \textbf{77.8 / 25.4} & \textbf{87.6 / 30.7} & \textbf{86.3 / 25.6} & 56.9 / 45.3                    
\end{tabular}
}
\caption{The performance of patches trained using the ensemble technique and evaluated on different YOLO versions.
$\text{Ens}_1$:YOLOv4 + YOLOv3; 
$\text{Ens}_2$:YOLOv5 + YOLOv4 + YOLOv3.}
\label{tab:ens}
\end{table}

\noindent\textbf{Different target classes.}
We examined our attack on two different target classes: bus and truck (using the YOLOv5s model and v1 frames).
Our evaluation shows that the attack was able to produce an efficient patch for both the target classes. 
For the bus target class, the patch changes the detection of 94.6\% of the cars (with a 12.2\% double detection rate). 
For the truck target class, the patch changes the detection of 88.9\% of the cars (with a 9.4\% double detection rate).

\subsection{Results: Physical Experiments}
\noindent\textbf{Effectiveness of our patch.}
To evaluate our patch in the physical domain, we use two toy cars -- red and white -- that YOLO correctly detects as car objects. 
We printed the patch on a piece of paper, placed it on the cars' hoods and filmed each of the cars driving towards the camera (a remote control was used to drive the cars).
Each video in the physical domain experiments contained approximately 80 frames (3 to 4 seconds worth of video).
In addition, similar to the evaluation in the digital domain, we printed a baseline patch with random pixel values that was the same size and shape of our adversarial patch.
Our patch's success rate on the toy cars was 95.9\% (with 7.5\% double detections), while the random baseline patch had a success rate of 3.7\% (and no double detections).
These results indicate that the efficiency of our physical patch is very similar to that of the digital patch and is even slightly better. 
We attribute this to the fact that in the real world the projection (placement) of the patch is always accurate and on the exact location on the car. \\

\noindent\textbf{Different car parts and shapes.}
To test our patch's robustness in the physical domain, we performed two types of experiments: (1) placing the patch on different parts of the toy car, and (2) changing our patch's shape to a different shape than it was originally trained on.
For the first type, we placed the patch on the car's trunk, when the car is driving away from the camera, and on the car's door, when filming the car from the side.
The success rate of our patch when placed on the car's trunk was 83\% (with 20.8\% double detections), and on the car's door it was 94.2\% (with 33.3\% double detections), \ie, the success rate remains high but at the same time the percentage of double detections increases.\\
For the second type, we changed the shape of the using a resizing function (after the training process), changing it to the following shapes: a wide rectangle shape and a narrow/slim rectangle shape.
The success rate of the stretched patch was 93.6\% (with 21.8\% double detections), and for the shrunken patch it was 72.6\% (with 15.3\% double detections).
Although the attack was successful when the shape of the trained patch was changed, the success rate 
 decreased while the double detection rate increased.
From these results, we can conclude that in the phyisical domain the patch should be implemented under the same conditions in which it was trained. 

\subsection{Results: Ablation Study}

\noindent\textbf{Finding relevant candidates using IoU.}
To demonstrate the contribution of our approach for finding relevant candidates in the attacked frame based on their IoU value, we trained a patch by using candidates that were identified in the attacked frame by the OD (similar to~\cite{thys2019fooling}). 
Our results show a significant decrease in the patch's performance. 
While the patch crafted using our method was able to change the detection of 94.6\% of the cars (with 12.2\% double detections), this patch changes the detection of only 55.3\% of the cars (with 42.6\% double detections).\\

\noindent\textbf{Projection function.}
To demonstrate the contribution of our projection function, we trained a patch employing the projection approach used by Thys \etal~\cite{thys2019fooling}. 
This approach places the patch in the center of the car's bounding box without adjusting the patch to the hood shape and without considering the car's relative location within the frame.
We tested the patch in the physical domain by placing the it on the toy cars' hoods and roofs, as well as on the toy cars' windshield which is the center of the cars.
The success rate (double detection) of the attack when using this patch are 68.6\% (13.4\%), 63.3\% (18.6\%), and 70.1\% (13.2\%) when placing the patch on the hood, roof, and windshield respectively.
These results clearly indicate that our projection function contributes significantly to the attack's success rate.

%% file: sections/07_conclusion.tex
\section{\label{sec:conclusion}Conclusion}

In this paper, we presented a universal targeted label-switch patch that changes an OD's detection of car on which it is applied, from the correct target class (car) to a bus. 
Our attack relies on several novel techniques aimed at optimizing the patch's performance: finding relevant candidates,  projecting the patch on the car in a realistic way, and a new loss function.  
The patch crafted by our attack is effective from different view angles and distances and therefore, can be used in the smart traffic systems use case, where the car is within range of a surveillance camera for a period of time, while driving on the street.
In future work, we plan to focus on developing a countermeasure capable of identifying adversarial patches on objects in real time.